\documentclass[10pt,twocolumn,letterpaper]{article}

\usepackage{cvpr}              

\usepackage{graphicx}
\usepackage{amsmath}
\usepackage{amssymb}
\usepackage{booktabs}

\usepackage[pagebackref,breaklinks,colorlinks]{hyperref}

\usepackage[capitalize]{cleveref}
\crefname{section}{Sec.}{Secs.}
\Crefname{section}{Section}{Sections}
\Crefname{table}{Table}{Tables}
\crefname{table}{Tab.}{Tabs.}

\def\cvprPaperID{*****} 
\def\confName{CVPR}
\def\confYear{2022}

\begin{document}

\title{A Discriminative Channel Diversification Network for Image Classification}

\author{Krushi Patel\\
Department of Electrical Engineering and Computer Science\\
University of Kansas\\
Lawrence, KS, USA, 66045\\
{\tt\small krushi92@ku.edu}
\and
Guanghui Wang\\
Department of Computer Science\\
Ryerson University \\
Toronto, ON, Canada M5B 2K3\\
{\tt\small wangcs@ryerson.ca}
}
\maketitle


\begin{abstract}
Channel attention mechanisms in convolutional neural networks have been proven to be effective in various computer vision tasks. However, the performance improvement comes with additional model complexity and computation cost. In this paper, we propose a light-weight and effective attention module, called channel diversification block, to enhance the global context by establishing the channel relationship at the global level. Unlike other channel attention mechanisms, the proposed module focuses on the most discriminative features by giving more attention to the spatially distinguishable channels while taking account of the channel activation. Different from other attention models that plugin the module in between several intermediate layers, the proposed module is embedded at the end of the backbone networks, making it easy to implement. Extensive experiments on CIFAR-10, SVHN, and Tiny-ImageNet datasets demonstrate that the proposed module improves the performance of the baseline networks by a margin of 3\% on average. 
\end{abstract}

\section{Introduction}

Deep convolutional neural networks (CNN) have become a dominant approach to solve a wide range of computer vision tasks, including image classification ~\cite{cen2021deep}, object detection ~\cite{ma2020mdfn}, semantic segmentation ~\cite{he2021sosd}, recognition ~\cite{sajid2021parallel}, image translation ~\cite{xu2019toward}. Inspired by the tremendous success of AlexNet ~\cite{krizhevsky2012imagenet} in image classification, many researchers have developed different network structures to boost the performance of deep CNN ~\cite{patel2021enhanced}. In recent years, the superior performance obtained by squeeze and excitation networks
~\cite{hu2018squeeze} has attracted many researchers to incorporate a channel attention mechanism in the convolutional neural networks.  

\begin{figure}[!t]
\centering
\includegraphics[width=0.9\linewidth]{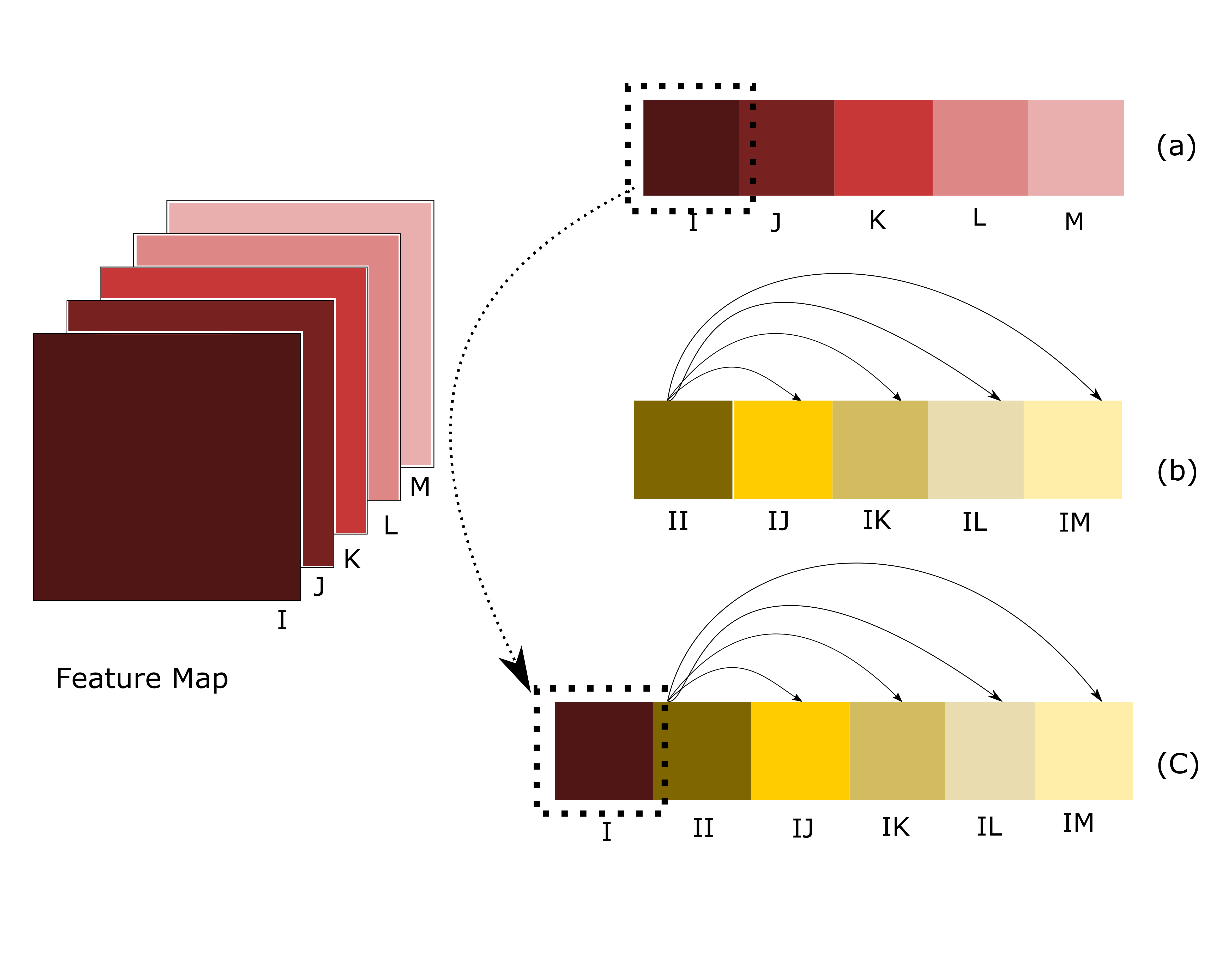}
\caption{Global relationship operation: (a) Features generated by global average pooling that demonstrates the overall significance of each channel; (b) features generated using the modified global attention pooling which represents how each channel is dissimilar with the given channel. Here we only display the relationship of one channel for the illustration purpose; (c) fusing of the features generated by (a) and (b) using concatenation operation.}
\label{fig:figure1}
\end{figure} 

Previous channel attention networks ~\cite{wang2018non,cao2019gcnet,gao2020channel,fu2019dual} utilize the collection of global information by calculating pair-wise relation between channels but overlook the significance of single channel information. 
We believe that the effect of the single channel's overall activation on global pairwise relation could exploit global information of channels well, without adding more extra parameters, unlike ABN ~\cite{fukui2019attention}.   

In general, each channel in the feature map focuses on a specific part of the image.  A natural tendency of the convolutional neural network is to focus only on a few class specified dominant channels, which limits the set of cues to classify the image. To alleviate this problem, we propose a novel channel attention mechanism, called channel diversification module, to force the network to learn more diverse and significant features by exploiting both the given channel's overall activation and pair-wise channel relationship as illustrated in Figure ~\ref{fig:figure1}. It shows that the significance of each channel is calculated by the global average pooling and to focus more on diverse features, global attention pooling is used,  which are fused together using concatenation.

\begin{figure*}[!t]  
\centering
\begin{tabular}{ccc}
 \includegraphics[width=0.25\linewidth]{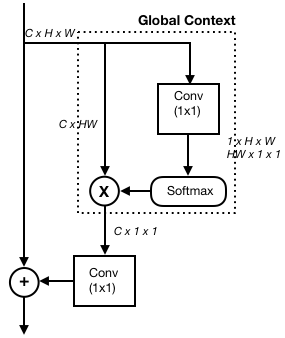} &\includegraphics[width=0.23\linewidth]{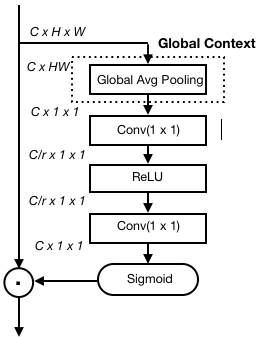} & \includegraphics[width=0.26\linewidth]{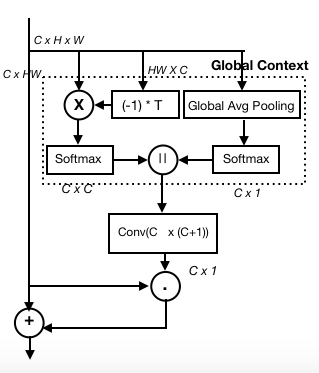} \\[-6pt]
(a) & (b) & (c) \\
\end{tabular}
\caption{Comparison of different attention based modules: (a) Simplified Non-local network; (b) SE block; and (c) channel diversification block. Where $C\times H\times W$ is the feature map dimension, $``+"$ denotes the broadcast elements-wise addition, $X$ denotes the matrix multiplication, $``."$ represents the broadcast elementwise multiplication, {{$``||"$ represents the concatenation operation, and $(-1)*T$ represents the multiplication of matrix with its transpose and $(-1).$}}}
\label{fig:figure2}

\end{figure*}

The proposed module can be considered as a combination of the simplified non-local block ~\cite{cao2019gcnet} and SE-block ~\cite{hu2018squeeze}. 
It concatenates the features generated by the global context modeling from both networks and performs a transformation without using MLP, unlike SE-block. Figure ~\ref{fig:figure2} shows the block diagrams for all three networks to illustrate the difference among them. Our module adopts the part which is highlighted using the dotted box in Figure~\ref{fig:figure2} (a) and (b) and concatenated the features generated by those parts as shown in Figure~  \ref{fig:figure2} (c).

The channel diversification module explores the input feature map from the last convolutional layer of the backbone network as the input and produces a C-dimensional feature vector using global average pooling, which represents the significance of each channel for a specific class. At the same time, it also generates a channel relationship matrix, which specifies how the given channel is distinct from other channels. After that, both the channel relation matrix and feature vector are stacked together, and convolution is applied to generate the weighted score for each channel.

The proposed channel diversification block focuses on spatially distinguished channels while considering the significance of each channel. It penalizes the most dominant channels and diverts the attention to different channels that are spatially distinguished from the given channel as well as have a large average activation. Therefore, our module only focuses on the channels which are diverse and significant enough at the same time. The new contributions of this work are summarized as follows:

\begin{itemize}
    \item The paper proposes a novel channel diversification block that makes the convolutional neural networks to focus on significant and diverse channels by establishing the relationship between the local (single channel) and global information channels.
    \item The proposed channel diversification block can be easily plugged in before the output layer of any baseline network to improve the performance of the baseline network by adding only a few extra trainable parameters and GFLOPs.
\end{itemize}

Extensive experiments have been conducted on CIFAR-10 ~\cite{cifar10}, CIFAR-100 ~\cite{cifar}, SVHN ~\cite{netzer2011reading}, and the Tiny-ImageNet dataset with various baseline networks, including VGGNet ~\cite{simonyan2014very}, ResNet ~\cite{he2016deep}, Wide-ResNet ~\cite{zagoruyko2016wide}, ResNext ~\cite{xie2017aggregated}, and DenseNet ~\cite{huang2017densely}. The results demonstrate that the proposed module
outperforms the baseline models and achieves competing performance with the different attention-based model while adding less computation cost.  The source code of the proposed module can be downloaded from the link: https://github.com/rucv/ChannelDiversification 


\section{Related Work}
There has been a rapid evolution in the field of image classification since the publish of Alexnet ~\cite{krizhevsky2012imagenet}, which achieves a record-breaking image classification accuracy. After that, researchers have been focused more and more on deep learning-based approaches for image classification tasks ~\cite{li2021sgnet}. VGG-Net ~\cite{simonyan2014very} and GoogleNet ~\cite{szegedy2015going} introduced a block-based architecture and proved that a deeper model could significantly improve the classification accuracy. ResNet ~\cite{he2016deep} proposed a skip connection based residual module to solve the vanishing gradient problem in deep models. ResNext ~\cite{xie2017aggregated} and Xception~ \cite{chollet2017xception} employed multi-branch architecture to increase the cardinality. Our proposed network employs the above-mentioned networks as a baseline and integrates the channel diversification block at the end.

The attention mechanism has proven to be very effective in various computer vision and natural language processing tasks.  They have been widely used in sequential models. In computer vision, it started after SE-Net's large performance gain using a channel attention mechanism ~\cite{hu2018squeeze}. Inspired by this idea, the residual attention network ~\cite{wang2017residual} introduced bottom-up and top-down feed-forward structure in
the attention mechanism, and CBAM ~\cite{woo2018cbam} used both max and average pooling to aggregate the features as well as compute spatial attention using 2D convolution.  Non-Local networks ~\cite{wang2018non} compute spatial attention by taking into account features from all
spatial positions. GC-Net ~\cite{cao2019gcnet} combines the non-local ~\cite{wang2018non} and SE-block ~\cite{hu2018squeeze} to produce a light-weight attention module. Our method utilizes the SE-block and non-local block in a new way.

Most previous attention mechanisms refine the intermediate features; we propose to apply the attention module at the end before the output layer like attention pooling ~\cite{girdhar2017attentional}, SOAL ~\cite{kim2020spatially}, ABN ~\cite{fukui2019attention}, AG-CNN ~\cite{guan2020thorax}, AGNN ~\cite{zou2020attention} and channel interaction network ~\cite{gao2020channel}. Attention pooling and SOAL computes spatial attention. The channel interaction network is similar to ours but it does not consider the single channel significance. ABN uses CAM ~\cite{zhou2016learning} generated attention weight to focus on a specific region of the image. AGNN used both self-attention and CAM-based attention and combined attentive features with LSTM. AG-CNN introduced threshold-based attention to generate a binary mask, which is further used to crop a global image to extract a significant local region.

\section{Proposed Method}
In this section, we describe the proposed channel diversification block in detail. Our proposed block is a combination of global features generated by simplified non-local block and SE-block. We first re-visit both blocks and represent which feature operation we adopted from them one by one, followed by a detailed explanation of the channel diversification network.
\begin{figure*}[!t]
\centering
\begin{tabular}{c}
 \includegraphics[width=0.7\linewidth]{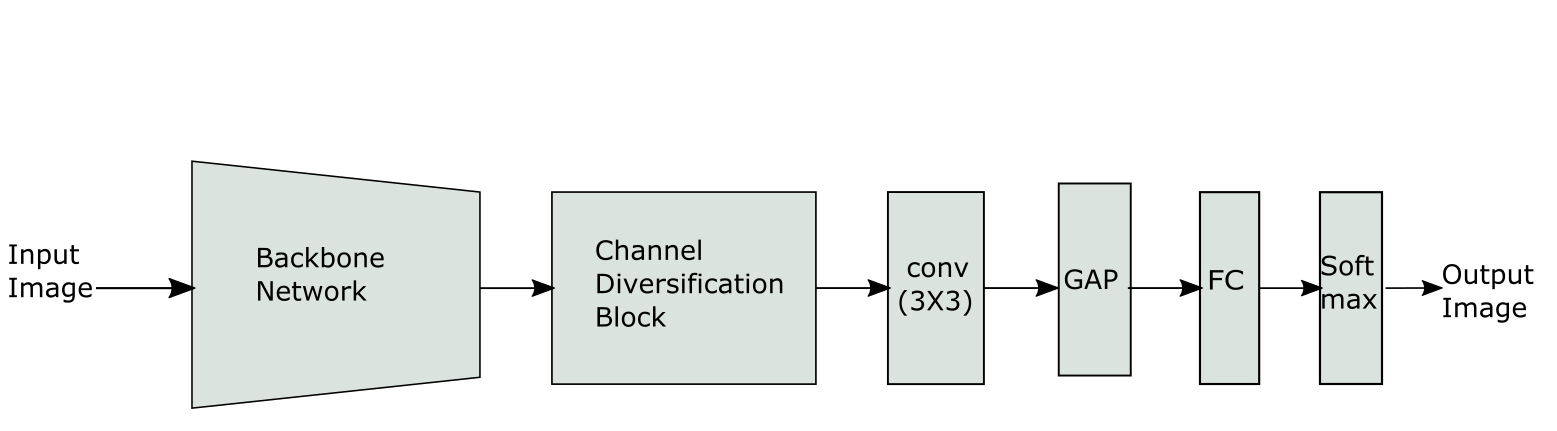} \\[-6pt]
(a) \\[-6pt]
\includegraphics[width=130mm,height=60mm]{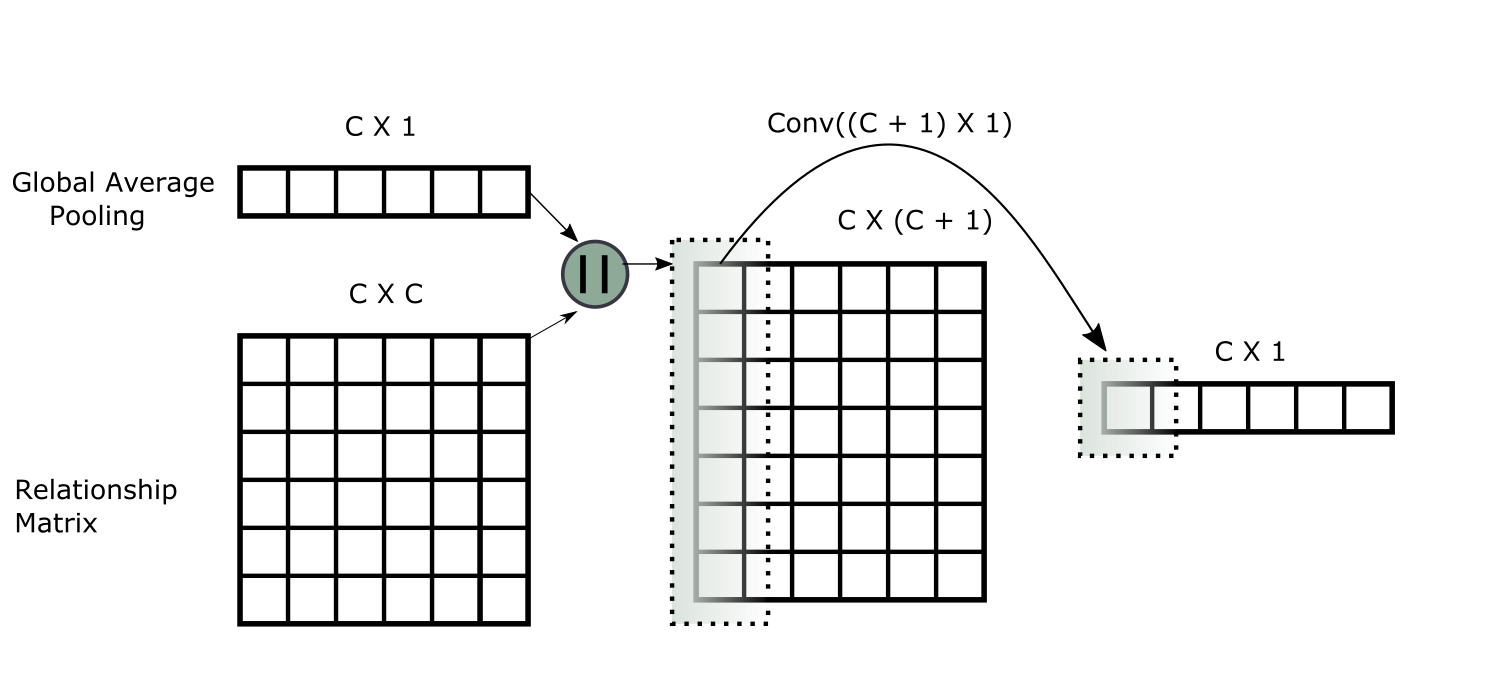} \\[-6pt]
(b)\\

\end{tabular}
\caption{(a) The overall architecture of the classification network with the channel diversification network. (b) The architecture of the channel diversification module. The features generated by the global average pooling and the modified global attention pooling are added elementwisely and multiplied by the original feature map. The dimension of the features is displayed on the top of each feature.} 
\label{fig:figure3}

\end{figure*}

\subsection{Revisit Simplified Non-Local Block:}
Simplified non-local block enhances the features of a given position by aggregating feature information of other remaining positions. It can be formulated as ~\cite{cao2019gcnet}: 


\begin{equation}
 z_{i} = x_{i} + W_{v}(\sum_{j=1}^{N_{p}}\frac{e^{W_{k}x_{j}}}{\sum_{m=1}^{N_{p}}e^{W_{k}x_{m}}}x_{j})
\end{equation}
Where $x$ is the one instance of the feature map, $N_{p} = W \times H$, where $W$ and $H$ is the width and height, and $W_{v}$ and $W_{k}$ is the linear transformation. From the above equation, we adopted only global attention pooling, and instead of calculating spatial relationship, we calculate the channel-wise relation ~\cite{gao2020channel} ~\cite{fu2019dual}, which can be formulated as:
\begin{equation}
\label{eq:ch}
\alpha_{ij} = \frac{e^{-x_{i}x_{j}}}{\sum_{m=1}^{C}e^{-x_{i}x_{m}}}
\end{equation}
Where C stands for the number of channels. $\alpha_{ij}$ represents the relationship of the $i^{th}$ channel to $j^{th}$ channel. It gives more weights to dissimilar channels and smaller weights to the most correlated channels.





\subsection{Revisit Squeeze and Excitation Block:}
The Squeeze and Excitation block extracts the global features and applies transformation using MLP, which selectively emphasizes informative features and suppresses less useful ones. It consists of two parts: (1) global average pooling for global context modeling and (2) series of $1 \times 1$ convolution layer followed by non-linear transformation. SE-Net can be formulated as ~\cite{hu2018squeeze}:


\begin{equation}
Z = \sigma(W_{2} \delta(W_{1}\alpha))
\end{equation}
Where $\alpha$ represents the global average pooling. From the SE block, we adopt the global average pooling part, which can be formulated as below.
\begin{equation}
\label{eq:avg}
\alpha_{i} =  \frac{1} {H \times W}\sum_{k=1, l=1}^{W, H}X_{k,l}
\end{equation}
Where W and H are the width and height of the feature maps, respectively. $ \alpha_{i}$ represents the average activation of the $i^{th}$ channel, which indicates the importance of each channel.

\subsection{Channel Diversification Network}


In the channel diversification block, we fuse Eq. \eqref{eq:ch} and Eq. \eqref{eq:avg} adopted from the simplified non-local block and the squeeze and excitation block, respectively. Features generated by both equations are $\alpha_{ij}$, and $\alpha_{i}$ represents the relationship between one channel to another and the overall importance of the channel, respectively, which are fused using concatenation operation as described in Figure ~\ref{fig:figure1}.

Our channel diversification block takes $X \in \mathbb {R} ^ {C \times H \times W}$ feature map and apply global average pooling, which
generates $C$ dimensional feature vector, $A \in \mathbb {R} ^ {C \times 1}$

\begin{equation}
A_{c} =  \frac{1} {H \times W}\sum_{k=1, l=1}^{W, H}X_{k,l}
\end{equation}

Softmax normalization $A = Softmax(A)$ is applied to the output of the global average pooling of all channels of the feature map, which represents the significance of each channel. At the same time, the feature map is applied to channel-wise simplified non-local block as formulated in Eq. \eqref{eq:ch} and produces a channel relation matrix $J \in \mathbb {R} ^ {C \times C}$.


\begin{equation}
J = -X.X^{T}
\end{equation}

After that, channel-wise softmax normalization is applied to the produced channel relation $J = Softmax(J)$, which indicates how dissimilar the given channel is to other remaining channels. Both the normalized features generated by the global average pooling and channel relationship matrix are then concatenated to yield the feature $Y \in \mathbb {R} ^ {C \times C+1}$

\begin{equation}
Y = A \ \Vert \ J
\end{equation}

We then apply a 2D transformation to this concatenated output using the convolution of size $1 \times (C+1)$ to enhance the global context by establishing the relationship between the single channel activation and global channel relationship and 
produce the weighted feature vector  $Y \in \mathbb {R} ^ {C \times 1}$

\begin{equation}
Y = f(Y)
\end{equation}
and the resultant attention vector is then multiplied and added to the original feature map. 

\begin{equation}
X = X \otimes Y + X
\end{equation}

After applying attention, the generated feature map is passed through one convolution layer, followed by the classification layer. 
The above-mentioned channel diversification block can be plugged into any classification network, as shown in Figure~\ref{fig:figure3} (a). The input image is first passed through a backbone network to generate the feature map $X = [x_{1}, x_{2}, ..., x_{c}]$, where $ X \in \mathbb {R} ^ {C \times H \times W}. $ This generated feature is passed through the channel diversification block, which forces the network to focus more on diverse features using channel attention pooling and significant features using global average pooling. The detail of the channel diversification block has been illustrated in Figure~\ref{fig:figure3} (b). In which, we can see that, average pooling generates the feature vector of size $ C \times 1 $ and channel-wise attention pooling generates a feature of size $C\times C$, are normalized and concatenated followed by linear transformation using convolution operation with kernel size $1 \times (C+1)$, which produces $C \times 1$ dimension features.

\section{Experiments}

We evaluate the channel diversification network on CIFAR-10, CIFAR-100, SVHN, and Tiny-ImageNet datasets and compare the performance with respect to various baseline networks.

\subsection{Datasets}
We evaluate the proposed approach on the following four publicly available benchmarks.
 
{\bf CIFAR-10:}
CIFAR-10 dataset consists a total of 60,000 images of size $32 \times 32$ and 10 classes, with 6,000 images per class. There are a total of 50,000 training and 10,000 testing images.

{\bf CIFAR-100:}
Similar to CIFAR-10, it consists of a total of 60,000 images of size $32 \times 32$, but has a total of 100 classes, with 600 images per class. There are a total of 50,000 training and 10,000 testing images.

{\bf SVHN:}
SVHN dataset consists of 604,388 training image(train: 73,257 and extra: 53,131) and 26,032 testing image of size $32 \times 32$.  It categorizes the images into 10 classes.

{\bf Tiny-ImageNet.}
The dataset contains 100,000 training images and 10,000 validation images of size $64 \times 64$. It categorizes the images into 200 classes.

\subsection{Training}

During training, we applied standard data-augmentations, which include zero-padding with 4-pixels on each side, randomly cropped them in the size of  $32 \times 32$ for CIFAR-10/100 and SVHN dataset, and $64 \times 64$ for the Tiny-ImageNet dataset, and randomly horizontally mirrored them. We trained the network using stochastic gradient descent (SGD) with momentum 0.9. CIFAR-10 and CIFAR-100 datasets are trained for 200 epochs with an initial learning rate of 0.1 and batch size 128 for all models except ResNext. For ResNext, we use the initial learning rate of 0.1 and batch size 64. The SVHN and Tiny-ImageNet datasets are trained for 50 and 200 epochs, respectively, with an initial learning rate of 0.1 and batch-size 128 and 256, respectively, for all models except ResNext. For ResNext, we use an initial learning rate of 0.01 and batch-size 64.

\begin{table}[!htb]
\begin{center}
\begin{tabular}{ | c | c | c |}
\hline
 &\multicolumn{1}{l|}{CIFAR-100} & \multicolumn{1}{l|}{CIFAR-10} \\
\hline
Model & Top- 1 & Top-1 \\
\hline
ResNet-110&73.12&93.57 \\
ResNet-110-SE&76.15&94.79\\
ResNet-110-SAOL&77.15&95.18\\
ResNet-110-ABN&77.15&95.09\\
ResNet-110-Ours &\textbf{77.50}&\textbf{95.60}\\
\hline
 WRN-16-8 &79.57&95.73\\
 WRN-16-8-SE &80.86&96.12 \\
 WRN-16-8-ours &\textbf{80.91}&\textbf{96.20}\\
\hline
WRN-28-10&80.13&95.83\\
WRN-28-10-SAOL&80.89&96.44\\
WRN-28-10-ABN&\textbf{81.88}&96.22\\
WRN-28-10-ours&81.86&\textbf{96.46}\\
\hline
ResNext&81.68&96.16 \\
ResNext-ABN&82.30&96.20 \\
ResNext-Ours&\textbf{83.02}&\textbf{96.43} \\
\hline
DenseNet&77.73 & 95.41\\
DenseNet-ABN&78.37 & 95.83 \\
DenseNet-SAOL&76.84&95.31 \\
DenseNet-Ours&\textbf{78.41}&\textbf{95.51} \\
\hline
VGG-16&72.18&92.64\\
VGG-16-ours&\textbf{74.67}&\textbf{94.29}\\
\hline
VGG-11&68.64&92.00\\
VGG-11-ours&\textbf{72.18}&\textbf{92.94}\\
\hline
\end{tabular}
\end{center}
\caption{Comparison of Top-1 accuracy on CIFAR-10 and CIFAR-100 datasets with various baseline models and attention based classification models. }\label{tab:table1}
\end{table}

\subsection{Accuracy on CIFAR-10/100 dataset}
Table~\ref{tab:table1} shows the top-1 accuracy on CIFAR-10 and CIFAR-100 datasets for various baseline models and our channel diversification network. The accuracy of the baseline models is taken from the original papers. 
Accuracy with $``*"$ indicates the result of re-implementation. 

The results indicate that our model consistently improves the performance of all baseline networks: ResNet-110, WRN-16-8,  WRN-28-10, ResNext,  DenseNet, VGG-16 and VGG-11 by 4.38\%, 1.34\%, 1.73\%, 1.34\%, 0.68\%, 2.55\% and 3.54\% respectively on the CIFAR-100 and 2.03\%, 0.47\%, 0.63\%, 0.27\%, 0.10\%, 1.65\% and 0.94\% respectively on the CIFAR-10 dataset. The highest accuracy on the CIFAR-100 dataset is  83.02\% and CIFAR-10 is 96.46\% , achieved by our channel diversification block with the baseline networks, ResNext, and WRN-28-10, respectively.

In the case of VGG-Nets, we removed the last max-pooling layer and classification layer and replaced those with the channel diversification block followed by a convolution layer. This modification further reduces the number of parameters and GFLOPs compare to the baseline VGG, as shown in Table 2.

From the result, it can be seen that our model outperforms the state-of-the-art SAOL network for all baseline networks.  In the case of ABN, our model outperforms all baseline networks except WRN-28-10 for CIFAR-100 and DenseNet for CIFAR-10. The accuracy difference between ABN and our model for WRN-28-10 and DenseNet is very small, around 0.02\% and 0.32\% on CIFAR-100 and CIFAR-10, respectively.

\begin{table}[tb]
\centering
\begin{tabular}{|c | c  c  c |}
\hline
Model& GFLOPS& Parameters& Top-1 \\
\hline
ResNet-110-SE&\textbf{0.21}&1.89M&76.15 \\
ResNet-110-ABN&0.60&3.06M&77.15\\
ResNet-110-Ours &\textbf{0.21}&\textbf{1.76M}&\textbf{77.20} \\
\hline
 WRN-16-8-SE &2.01 &11.2M&80.86 \\
 WRN-16-8-ours &\textbf{1.01} &\textbf{11.0M} &\textbf{80.91} \\
\hline
WRN-28-10-ABN&12.4 & 64.48M&\textbf{81.88} \\
WRN-28-10-ours&\textbf{5.46}&\textbf{36.5M}&81.86\\
\hline
ResNext-ABN& \textbf{0.25}& 120.32M&82.30 \\
ResNext-Ours& 0.69&\textbf{ 77.79M}&\textbf{83.02}  \\
\hline
DenseNet-ABN&\textbf{0.32} &\textbf{1.12M}&78.37 \\
DenseNet-Ours& 0.37&1.83M&\textbf{78.41} \\
\hline
VGG-16&0.33&34.02M&72.18\\
VGG-16-ours&\textbf{0.32}&\textbf{17.4M}&\textbf{74.67}\\
\hline
VGG-11&0.17&28.52M&68.64\\
VGG-11-ours&\textbf{0.16}&\textbf{11.6M}&\textbf{72.18}\\
\hline
\end{tabular}
\caption{Comparison of Top-1 accuracy, number of parameters, and GFLOPs on the CIFAR-100 dataset with other attention-based classification models. }
\label{tab:table2}
\end{table}

It is also evident from Table~\ref{tab:table2} that our model requires less computation and has a fewer number of parameters compared to ABN. For example, ResNet-110 with ABN requires 0.6 GFlops and 3.06M parameters, while ResNet-110 with our channel diversification block requires only 0.21 GFLOPs and 1.76M parameters. The comparison of accuracy vs GFLOPs, and the number of parameters for different baseline network are shown in Table~\ref{tab:table2}.

\begin{table}[!tb]
\begin{center}
\begin{tabular}{|c| c  | }
\hline
Model&Top-1\\
\hline
 ResNet-110&97.82 \\
ResNet-ABN&98.14\\
 ResNet-110-Ours &\textbf{98.15}\\
\hline
WRN-28-10& 97.58\\
WRN-28-10-ABN& 97.76\\
WRN-28-10-ours&\textbf{98.20}\\
\hline
ResNext&97.84\\
ResNext-ABN&97.99\\
ResNext-Ours&\textbf{98.00} \\
\hline
DenseNet& 97.93\\
DenseNet-ABN&97.99\\
DenseNet-Ours&\textbf{97.99}\\
\hline
\end{tabular}
\end{center}
\caption{Comparison of Top-1 accuracy on SVHN dataset with different baseline models and ABN. }
\label{tab:table3}
\end{table}

\subsection{SVHN Accuracy}
We evaluate the performance of our model on the SVHN dataset. The top-1 accuracy on SVHN has been shown in Table~\ref{tab:table3}, in a similar manner as stated in the CIFAR-10/100 accuracy section. From the result, it is clear that our model outperforms all the baseline networks: ResNet-110,  WRN-28-10, ResNext, and DenseNet on the SVHN dataset in terms of accuracy. The results also show that our model achieves competing performance with ABN. For some baseline models, it exceeds the performance of ABN. For example, the baseline networks ResNet-110 and WRN-28-10 outperform the ABN by 0.33\% and 0.62\%  on SVHN dataset, respectively.

\subsection{Tiny-ImageNet Accuracy}
We also evaluate the performance of our model on the Tiny-ImageNet dataset. The top-1 accuracy on this dataset is shown in Table~\ref{tab:table4}, from which we can see that the channel diversification block significantly improves the performance of all baseline networks: ResNet-110, DenseNet, Wide-ResNet, and  ResNext by 0.71\%, 3.03\%, 2.01\%, and 2.05\%, respectively. The highest accuracy we achieved on the Tiny-ImageNet dataset is 70.28\% by ResNext(ours). In this experiment, the baseline accuracy of various models was obtained by re-implementing all models following the published papers.

\begin{table}[!tb]
\begin{center}
\begin{tabular}{|l|c|}
\hline
Model& Accuracy \\
\hline
ResNet-110 & 62.56* \\
ResNet-110(ours) &\textbf{63.27}\\
\hline
DenseNet & 60.00* \\
DenseNet(ours) &\textbf{63.03} \\
\hline
Wide-ResNet & 65.99*\\
Wide-ResNet(ours)&\textbf{68.00}\\
\hline
ResNext& 68.23*\\
ResNext(Ours)&\textbf{70.28}\\
\hline
\end{tabular}
\end{center}
\caption{Comparison of Top-1 accuracy of various baseline models with ours. Here, $``*"$ indicates the re-implementation accuracy.}
\label{tab:table4}
\end{table}

\begin{table*}[!tb]
\begin{center}
\begin{tabular}{|c | c| c|}
\hline
Model &Accuracy& Parameters\\
\hline
ResNet-110(Baseline)&73.12&1.70M\\
ResNet-110 (with only Global Avg Pooling)&76.56&1.70M \\
ResNet-110 (with only Attention Pooling(using negative correlation))&77.20&1.70M \\
ResNet-110 (SE)\cite{hu2018squeeze}& 76.15&1.89M\\
ResNet-110 (GC (using positive correlation))& 74.65&1.92M \\
ResNet-110 fusion(apply after each residual block)&76.93 & 1.84M\\
ResNet-110 fusion(using positive correlation) &76.13 &1.76M \\
ResNet-110 fusion(Ours) & \textbf{77.50} & \textbf{1.76M}\\
\hline
\end{tabular}
\end{center}
\caption{Comparison of top-1 accuracy on CIFAR-100 only using either the global average pooling, or the global attention pooling, or using both. We also show the individual accuracy of the attention model SE-Net and GC-Net.}
\label{tab:table5}
\end{table*}

\subsection{Visualizing Attention Maps}

\begin{figure}[!tb]  
\begin{center}
\begin{tabular}{cccc}
Original&ResNet-110& Ours \\
\includegraphics[width=0.23\linewidth]{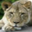}&\includegraphics[width=0.23\linewidth]{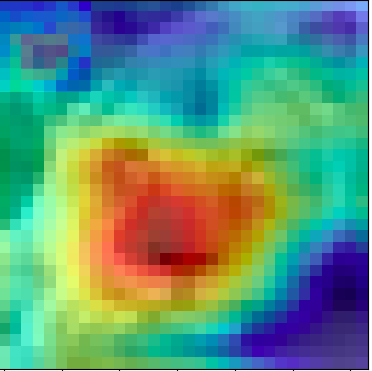} & \includegraphics[width=0.23\linewidth]{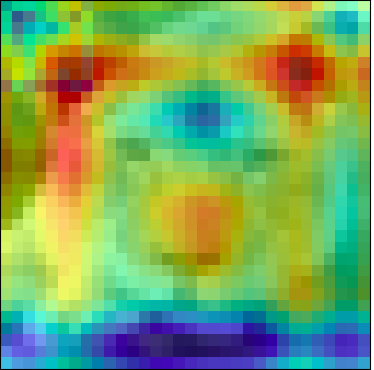} \\
Lion& & \\
\includegraphics[width=0.23\linewidth]{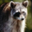} &\includegraphics[width=0.23\linewidth]{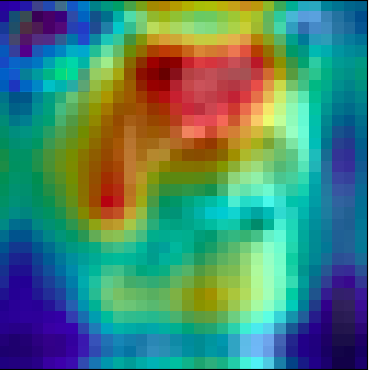} & \includegraphics[width=0.23\linewidth]{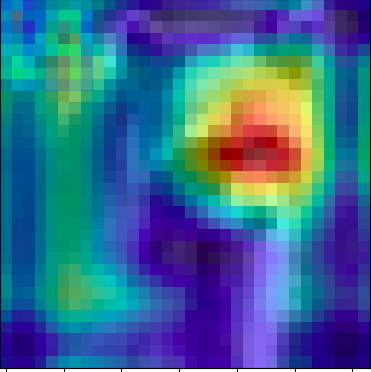} \\
Racoon& & \\
\includegraphics[width=0.23\linewidth]{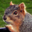} &\includegraphics[width=0.23\linewidth]{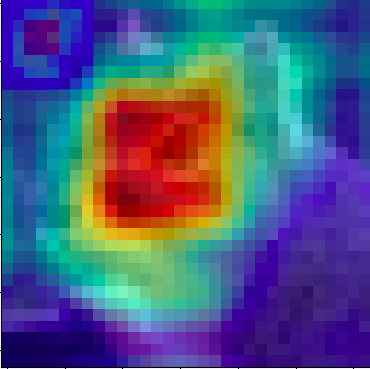} & \includegraphics[width=0.23\linewidth]{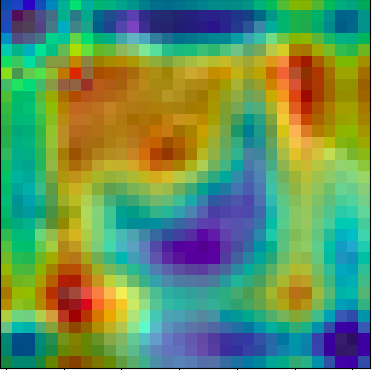} \\
Squirrel& & \\
\includegraphics[width=0.23\linewidth]{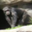} &\includegraphics[width=0.23\linewidth]{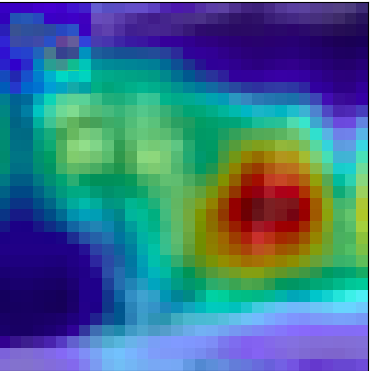} & \includegraphics[width=0.23\linewidth]{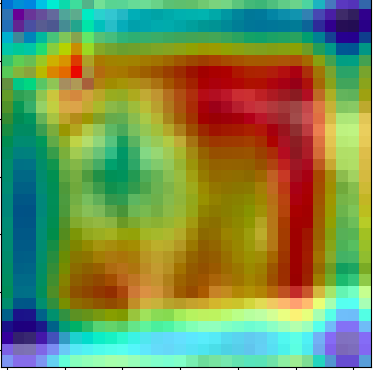} \\
Chimpanzee& & \\
\includegraphics[width=0.23\linewidth]{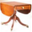} &\includegraphics[width=0.23\linewidth]{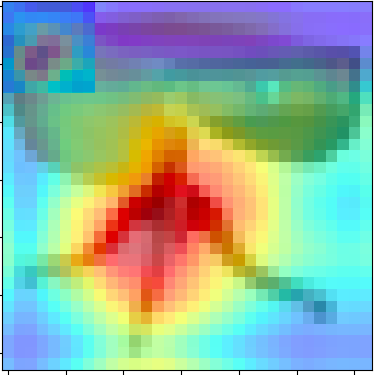} & \includegraphics[width=0.23\linewidth]{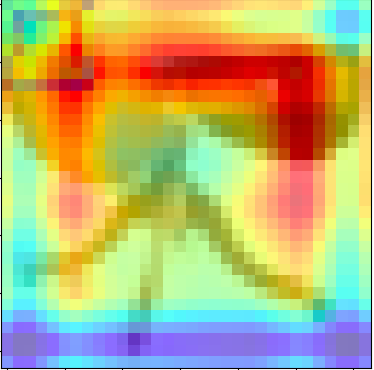} \\
Table& & \\
(a)&(b)&(c)\\[-6pt]
\end{tabular}
\end{center}
\caption{Comparison of the attention maps generated by the baseline model and the channel diversification network. (a) Original images; (b) attention maps generated by the baseline model; and (c) attention map generated by our channel diversification block. }
\label{fig:figure4}

\end{figure}

We compare the attention maps generated by the baseline network and our channel diversification network using CAM ~\cite{zhou2016learning}. All the attention maps displayed in Figure~\ref{fig:figure4} are from the images which could not be correctly classified by the baseline network. From the attention map, we can see the reason behind the misclassification. It gives attention to very common features, such as in the case of Lion, Racoon, and Squirrel image, it focuses on the nose, head, and face part, respectively, whereas our channel diversification network diverts the attention of the model to more diverse and significant features. For example, in the case of the Lion image, our model focuses on ears and nose; for the Racoon image, it gives attention to the face and body; and for the Squirrel image, the model focuses on ears and eyes, which are very important features classification. Similarly, in the case of the Chimpanzee and Table image, the baseline model focuses on unimportant and unrelated features, whereas our model focuses on the right place. We can also see from Figure~\ref{fig:figure4} that our model did not focuses on insignificant background features, though it diverts the attention. 

\section{Ablation Study}


In this section, we report ablation experiments to demonstrate the effectiveness of fusing the global context modeling part of the SE block and simplified the non-local block. We train ResNet-110 on CIFAR-100 by either using only the global average pooling from the SE block or attention pooling from the simplified non-local block, followed by a transformation that includes convolution with kernel size $1\times1$ and $C \times 1$ respectively. Our result shows that fusing both of these context modelings and applying the transformation with one convolution layer of size $C \times (C+1)$ on fusion features clearly improves the performance. We also plugin SE-Net and Simplified Non-Local Block at the end of the backbone network. From Table~\ref{tab:table5}, it is evident that our model outperforms the SE-Net and the simplified non-local block individually on the CIFAR-100 dataset. In addition,  we conduct two comparative experiments by either using positive correlation in our module or inserting our module after each residual block, as shown in Table~\ref{tab:table5}. We can see from the experiments that their performance is inferior to the proposed approach. We also compare the number of parameters for each of the cases for ResNet-110 and show that our model has a comparatively low number of parameters while achieves high accuracy of 77.5\%.

\section{Conclusion}
We have proposed a channel diversification block in this paper. The proposed scheme can be embedded in any CNN-based baseline networks to make them concentrate more on the diverse and significant channels at the same time. We have extensively evaluated the performance of the channel diversification block for image classification tasks on various datasets. Our model outperforms all baseline networks and the attention-based classification models: SOAL, ABN, and SE-Net in terms of accuracy, the number of extra parameters, and GFLOPs. 

\section*{Acknowledgement}

This work was partly supported in part by the Natural Sciences and Engineering Research Council of Canada (NSERC) under grant no. RGPIN-2021-04244, the United States Department of Agriculture (USDA) under Grant no. 2019-67021-28996, and the National Institutes
of Health (NIH) under grant no. 1R03CA253212-01.

\bibliographystyle{ieee_fullname}
\bibliography{egbib}

\end{document}